%% file: preambulo.tex
\documentclass[sigconf]{acmart}

\usepackage{booktabs} 
\usepackage[ruled,linesnumbered,commentsnumbered,vlined,lined,boxed]{algorithm2e}
\usepackage{tikz}
\usepackage{multirow}
\usepackage{subcaption}
\usepackage{listings}

\usepackage{etoolbox}

\setcopyright{rightsretained}

\copyrightyear{2022}
\acmYear{2022}
\setcopyright{acmcopyright}\acmConference[GECCO '22]{Genetic and Evolutionary Computation Conference}{July 9--13, 2022}{Boston, MA, USA}
\acmBooktitle{Genetic and Evolutionary Computation Conference (GECCO '22), July 9--13, 2022, Boston, MA, USA}
\acmPrice{15.00}
\acmDOI{10.1145/3512290.3528695}
\acmISBN{978-1-4503-9237-2/22/07}

\begin{document}
\title{Transformation-Interaction-Rational Representation for Symbolic Regression}

\author{Fabrício Olivetti de França}
\affiliation{%
  \institution{Universidade Federal do ABC}
  \institution{Center for Mathematics, Computing and Cognition}
  \city{Santo André} 
  \state{SP}
  \country{Brazil}
}
\email{folivetti@ufabc.edu.br}


\begin{abstract}
  
   Symbolic Regression searches for a function form that approximates a dataset often using Genetic Programming. 
   Since there is usually no restriction to what form the function can have, Genetic Programming may return a hard to understand model due to non-linear function chaining or long expressions. A novel representation called Interaction-Transformation was recently proposed to alleviate this problem. In this representation, the function form is restricted to an affine combination of terms generated as the application of a single univariate function to the interaction of selected variables. This representation obtained competing solutions on standard benchmarks. Despite the initial success, a broader set of benchmarking functions revealed the limitations of the constrained representation.
   In this paper we propose an extension to this representation, called Transformation-Interaction-Rational representation that defines a new function form as the rational of two Interaction-Transformation functions. Additionally, the target variable can also be transformed with an univariate function. The main goal is to improve the approximation power while still constraining the overall complexity of the expression.
   We tested this representation with a standard Genetic Programming with crossover and mutation. The results show a great improvement when compared to its predecessor and a state-of-the-art performance for a large benchmark.
\end{abstract}



\maketitle

\input{conteudo}

\apptocmd{\sloppy}{\hbadness 10000\relax}{}{}

\balance

\bibliographystyle{ACM-Reference-Format}
\bibliography{referencias} 

\end{document}

%% file: conteudo.tex

\section{Introduction}~\label{sec:introduction}

Regression analysis is an important field that studies the relationship of measured quantities~\cite{kass1990nonlinear, harrell2017regression}. This has applications in a broad set of fields like economics, public health, engineering, etc~\cite{gelman2020regression}.

Most frequently, such analysis is performed by means of parametric approaches where we fix a certain function form following some assumptions and adjust its parameters to a given set of data points~\cite{kass1990nonlinear,harrell2017regression,gelman2020regression}.
Distinct from the parametric approach, \emph{Symbolic Regression}~\cite{koza1994genetic, langdon1999size, poli2008field} searches for the optimal function form altogether with the adjusted parameters corresponding to the best fit of a certain studied phenomena. This has the advantages that no prior assumptions are required (e.g., linearity, homoscedasticity, etc.). The evolutionary approach called \emph{Genetic Programming} (GP)~\cite{koza1994genetic, poli2008field} is often employed to solve this particular problem.

In the standard approach, the representation of a solution in GP is an expression tree data structure that is only constrained by the set of symbols allowed in the function nodes. The main advantage is that the there is no requirement of a prior study step, the search algorithm is in charge of finding the adequate function form. In practice, though, this not always works as expected since the search space can be hard to navigate as a small change in the expression can make up for a very different prediction behavior. 

A new representation was recently proposed in~\cite{SymTree,LightweightSymbolicRegressionWiththeIT} that constraints the search space to a specific pattern composed of an affine combination of non-linear transformed features. These features are a composition of a polynomial, representing the interactions between the original variables, and the application of an unary function, called transformation. This representation, called Interaction-Transformation, rendered competitive results when compared to modern symbolic regression approaches~\cite{2021srbench}. Its main advantage is the guarantee that some complicated function forms are not representable, such as those models containing non-linear function chaining. Also, the model parameters can be determined by using an ordinary least squares solver, removing this burden from the evolutionary search.

However, the imposed restrictions may keep the search algorithm from finding the correct solution, if the ground truth is not representable. In this paper we propose an extension to this representation, called Transformation-Interaction-Rational, that allows to represent a broader class of function forms while still limiting the search space to avoid overcomplicated function forms.

The remainder of this paper is organized as follows. In Section~\ref{sec:gp} we explain the Symbolic Regression problem with a brief detail of Genetic Programming and its common issues. Section~\ref{sec:related} gives a brief review of the current approaches proposed to alleviate such problems. In Section~\ref{sec:tir}, we start with a short explanation of the Interaction-Transformation followed by our proposed representation, Transformation-Interaction-Rational. Section~\ref{sec:evo} describes the implementation details of the evolutionary algorithm used to search for symbolic expressions. The experimentation method is described in Section~\ref{sec:method} followed by the results and discussions in Section~\ref{sec:results}. Finally, in Section~\ref{sec:conclusion}, we conclude this paper with a summary of the results and discussing some future steps.

\section{Genetic Programming}\label{sec:gp} 

Genetic Programming is an evolutionary search specialized in evolving computer programs~\cite{koza1994genetic, langdon1999size, poli2008field}. Specifically for Symbolic Regression, the computer program is an abstract syntax tree representing a mathematical expression that describes the observed relationship. This algorithm follows the traditional framework of Genetic Algorithm that repeats until convergence the cycle of selection, crossover, mutation, and reproduction.

A common crossover operator is the sub-tree crossover in which the algorithm chooses random subtrees from distinct parents and recombines them into new solutions. The mutation operator can be a choice of inserting, removing or replacing a random subtree, or a special case of replacing a node with another one of the same arity.

One issue with this procedure is that the tree can grow indefinitely without a proper control mechanism. This can lead to unnecessarily long expressions, that are harder to understand or contain unused subtrees (e.g., part of expression that evaluates to $+ 0$) also known as \emph{bloat}~\cite{dick2014bloat}. This problem can be alleviated by penalizing larger solutions or stipulating a maximum tree depth and prohibiting children solutions larger than this depth.

Another issue is the creation of nodes representing fixed model parameters or coefficients. A natural procedure to create this type of nodes is to draw a random value within a reasonable range. This creates a sub-problem within Symbolic Regression since, even if the algorithm finds the correct function form, it still has to find the correct coefficients. For example, let us suppose that the correct function form is $\sin{(c \cdot x)}$ with $c = 0.5$. If the algorithm finds the correct form but with $c = 5$, it will likely discard this expression due to a large approximation error.

Many authors proposed different solutions to this problem. Most noteworthy is the \emph{Linear Scaling}~\cite{keijzer2004scaled} that, for a given symbolic model $f(x)$, it fits a simple regression $a \cdot f(x) + b$ to adjust the scale and offset of the model. This simple procedure can help to maintain correct function forms throughout the generation with the expectation of eventually finding a better value for their coefficients. More recently, there has been some studies on applying non-linear optimization to adjust the scaling and offsets of inner nodes~\cite{kommenda2020parameter}, this particular approach has been proved to be successful in a thorough benchmark framework~\cite{burlacu2020operon}.

In this same line, some authors propose the combination of smaller models using an affine combination, this is called Multiple Regression GP~\cite{arnaldo2014multiple} and is best represented by FEAT~\cite{cava2018learning, LaCava:2019:SVO:3321707.3321776} and ITEA~\cite{ITEAforSymbolicRegression,ParametricStudyITEA,de2021interaction}, briefly explained in the next section.

\section{Related Research}\label{sec:related}

As mentioned in the previous section, the recent literature focuses on creating mechanisms that help to improve the exploration power of the genetic programming algorithm while reducing the influence or the need to search for model parameters.

\textbf{Feature Engineering Automation Tool (FEAT)}~\citep{cava2018learning,LaCava:2019:SVO:3321707.3321776} represents the regression model as the affine combination of smaller expression trees (sub-trees). The main idea is that the search problem is decomposed into smaller segments where the affine coefficients act as a sub-tree selection and scaling. The algorithm fits the coefficients using a gradient descent algorithm to find the nearest local optima.
In this algorithm, the initial population is formed of a trivial linear model and a set of small solutions created at random. These random solutions are created as an affine combination of expression trees with a limited depth. The main operators for this algorithm is the sub-tree crossover, that swaps sub-trees of two or more expressions; dimension crossover, that swaps two variable nodes between parents, and the mutations: node replacement, replacement of a sub-tree by a terminal node, sub-tree removal, insertion of a variable node, removal of a variable node. Besides the functions and operators commonly used in the literature, the authors also included functions often used as activation functions of Neural Network as well as boolean functions. 

\textbf{Interaction-Transformation Evolutionary Algorithm} \textbf{(IT--EA)} \cite{ITEAforSymbolicRegression,ParametricStudyITEA,de2021interaction} also describes an affine combination of smaller expressions (denoted as terms) but, differently from FEAT, it constrains the form of these expressions as the application of a single unary function, called transformation, to the product of powers of the original features, called interaction (more details in Sec.~\ref{sec:tir}). The algorithm is a mutation-based evolutionary search that applies a mutation to every solution in the population and selects the next generation using tournament selection. The mutation operators can either add or remove a new term, replace the transformation function, replace one of the exponents, combine the interactions between two terms. The results reported for a commonly used set of regression datasets~\cite{ITEAforSymbolicRegression} showed a competitive performance of ITEA when compared to FEAT and other SR algorithms. More recently, a more thoroughly benchmark~\cite{2021srbench} showed that in some situations this representation may limit the goodness-of-fit obtained by this algorithm. In this paper we propose an extension to this representation that alleviates this problem.

\textbf{Genetic programming with nonlinear least squares (GP NLS)}~\cite{kommenda2020parameter} is a recent modification of the original GP that when evaluating a solution it adds a scaling parameter to every node representing a variable and envelopes the expression tree with a linear scaling. The parameters are then determined by a nonlinear least squares method using the Jacobian of the expression. This is possible since the set of non-terminal have a well determined derivative that allows the use of automatic differentiation for the whole expression. The authors reported an outstanding result in a set of benchmark datasets when compared to other common approaches~\cite{kommenda2020parameter}. Its current optimized implementation, called Operon~\cite{burlacu2020operon} was used in a large benchmark of Symbolic Regression approaches and shown to outperform other SR and nonlinear regression approaches, on average, in both goodness-of-fit and runtime~\cite{2021srbench}.

\begin{table*}[t!]
    \centering
    \caption{Examples of mathematical functions and their corresponding representation as IT and TIR expressions, if any.}
    \label{tab:ITRepresentationExamples}
    \input{floats/texts/ITRepresentationExamples}
\end{table*}

\section{Transformation-Interaction-Rational}\label{sec:tir}

The \emph{Interaction-Transformation} representation (IT)~\cite{SymTree} constrains the function form of the mathematical model to those forms that can be represented as an affine combination of non-linear transformations applied to polynomial functions (interactions). The main idea comes from the observation that many engineering and physics equations can be described in this form.

Considering a tabular data set where each sample point has $d$ variables  $\mathbf{x} = (x_{1}, x_{2}, \cdots, x_{d})$. The regression model following the Interaction-Transformation (IT) representation has the form:

\begin{equation} \label{eq: ExprIT}
    f_{IT}(\mathbf{x, w}) = w_0 + \sum_{j = 1}^{m}{w_{j} \cdot (f_j \circ r_j) (\mathbf{x})},
\end{equation}
representing a model with $m$ terms where $\mathbf{w} \in \mathbb{R}^{m+1}$ are the coefficients of the affine combination, $f_j : \mathbb{R} \rightarrow \mathbb {R}$ is the $j$-th transformation function and $r_j : \mathbb{R}^d \rightarrow \mathbb {R}$ is the interaction function:

\begin{equation} \label{eq: term}
    r_j(\mathbf{x}) = \prod_{i = 1}^{d}{x_i^{k_{ij}}},
\end{equation}

\noindent where $k_{ij} \in \mathbb{Z}$ represents the exponents for each variable. Whenever we fix the values of $f_j$ and $r_j$, the expressions becomes a linear model with $\mathbf{w}$ as the coefficients, as such we can find their optimal values with an ordinary least squares method. This is convenient for the main search algorithm that only needs to find the optimal value of $m$, the values for the exponents and the functions of each term. Computationally, we can represent such expression as a list of tuples in the form $(\mathbf{k_j}, w_j, f_j)$ with the the $0$-th term represented as $(\mathbf{0}, w_0, id)$.

To illustrate the possibilities and limitations of this representation, in Table~\ref{tab:ITRepresentationExamples} we show four simple equations in which only one of them can be represented as an IT expression. The second column shows the same equation rewritten with a similar notation to an IT expression. Notice that, even though not every equation can be represented, this representation is still capable of finding good approximations~\cite{LightweightSymbolicRegressionWiththeIT,ITEAforSymbolicRegression,de2021interaction,kantor2021simulated}.

We now propose the Transformation-Interaction-Rational representation (TIR)~\footnote{Fun fact: Tir was the god of wisdom in ancient Armenia.} as an extension of the IT expressions. The main idea is that we combine two IT expressions similarly to a rational polynomial regression model~\cite{taavitsainen2010ridge, taavitsainen2013rational, moghaddam2017statistical} and apply an invertible function to the resulting value:

\begin{equation*}
    f_{TIR}(\mathbf{x, w_p, w_q}) = g\left(\frac{p(\mathbf{x, w_p})}{1 + q(\mathbf{x, w_q})}\right),
\end{equation*}
where $g : \mathbb{R} \rightarrow \mathbb{R}$ is an invertible  function, $p, q : \mathbb{R}^d \rightarrow \mathbb{R}$ are IT expressions exactly as defined in Eq.~\ref{eq: ExprIT} with $m_p > 0$ and $m_q \geq 0$ terms. 
With this function form we can still use an ordinary least squares method to adjust the coefficients $\mathbf{w_p, w_q}$ to the training data. Assuming $\mathbf{y_{tr}} = f_{TIR}(\mathbf{x_{tr}, w_p, w_q})$ a vector containing the values of the application of the TIR model to the training data $x_{tr}$, we can derive the following expression:

\begin{align*}
    y_{tr} &= g\left(\frac{p(\mathbf{x, w_p})}{1 + q(\mathbf{x, w_q})}\right) \\
    g^{-1}(y_{tr}) &= \frac{p(\mathbf{x, w_p})}{1 + q(\mathbf{x, w_q})} \\ 
    g^{-1}(y_{tr}) &= p(\mathbf{x, w_p}) - g^{-1}(y_{tr}) \cdot q(\mathbf{x, w_q}).
\end{align*}

We can adjust the coefficients using the same fitting algorithm (OLS) using a transformed training target by applying the inverse of the transformation function $g$. For that to work, it is required that the function $g$ is invertible and that $y_{tr}$ is within the domain of the inverted function when $g^{-1}$ is partial. Additionally, it is important that $p$ contains at least one term to avoid the trivial solution where $q(\mathbf{x, w_q}) = -1$.

To avoid the effect of the collinearity induced by the 
term $g^{-1}(y_{tr}) \cdot q(\mathbf{x, w_q})$, some authors suggest to apply a regularization factor~\cite{taavitsainen2010ridge, taavitsainen2013rational} or to fit using only a smaller subset of the data points~\cite{gaffke199630}. In some initial experiments, we did not observed any benefit in any of these approaches, thus this will be left as a future investigation.

Finally, the adjusted model should be evaluated with a validation set containing at least some distinct samples from the training data to avoid keeping overfitted models in the population.

In Table~\ref{tab:ITRepresentationExamples} we illustrate the equivalent form for three out of the four expressions used as an example. We can see that this new representation expands the search space represented by IT allowing us to represent more function forms while still keeping some simplicity. There are, of course, still some expressions that cannot be exactly represented by TIR but, as it was the case with IT, they still can be approximated.

\section{TIR Framework}\label{sec:evo}

To evaluate the potential of this representation we have implemented an evolutionary algorithm that searches for the best fitting TIR expression. This framework is written in Haskell and it is available at \url{https://github.com/folivetti/tir}, we also provide a scikit-learn compatible Python wrapper.

The framework uses three helper libraries for different steps of the evolutionary process: \emph{modal-interval}, \emph{srtree}, \emph{evolution}.

The \emph{modal-interval} library~\footnote{\url{https://github.com/folivetti/modal-interval}} is an implementation of the Modal Interval Arithmetic~\cite{goldsztejn2008modal}, an extension of the Interval Arithmetic~\cite{hickey2001interval}, capable of giving an estimate of the image of a function given the input domain ranges such as the true image is inside the estimated range. In TIR framework this is used to evaluate what functions $g^{-1}$ can be evaluated within the range of the training target and to remove any invalid terms of the IT expressions $p, q$. Notice that since any term of the IT expression contains at most a single occurrence of each variable, interval arithmetic returns an exact interval of the image of the function. To estimate the domains, we calculate the minimum and maximum values for each variable in the training data. The framework also allows the user to manually insert the corresponding domains if they are known \emph{a priori}.

The \emph{srtree} library~\footnote{\url{https://github.com/folivetti/srtree}} is responsible for managing and evaluating the expression trees. It supports optimized evaluation of vectorized data and the calculation of the derivative of any order for a given expression. This library is used to evaluate the validation data using the fitted expression, to generate a report of the properties of the generated model (i.e., monotonicity, etc.), convert the expressions into different formats (e.g., Python Numpy compatibility), and to apply a non-linear fitting algorithm (implemented but not used in this work).

Finally, the \emph{evolution} library~\footnote{\url{https://github.com/folivetti/evolution}} is a generic framework developed to help with the implementation of evolutionary algorithms. This library provides support for a high-level domain specific language and it automatically handles the search main loop. In short, one just needs to provide the implementation of the specific search operators (i.e., initialization, crossover and mutation) and a description of the desired search process so that the library handle the search. 



In this paper we are using the generational replacement with the application of a one-point crossover with probability $pc$ and using two parents selected by tournaments between two individuals, followed by a multi-mutation with probability $pm$.
Having defined the main loop of the evolutionary search, we need to determine the three specific operators: initialization, crossover, and mutation. 

\subsection{Initial Population}

Each solution is represented as a triple $(g, p, q)$ where $g$ is the invertible function, and $p, q$ are IT-expressions represented as a list of triples $[(w_i, f_i, [k_{ij}])]$. For any purpose, we can translate this representation into an expression tree and vice-versa.
The initial population is composed of random solutions limited by a user defined budget for the sum of the number of terms of the IT expressions $p$ and $q$. The procedure starts by first choosing a random invertible function $g$ from a set $G$ of invertible functions, then generating a random non-null IT expression $p$ and, finally, an IT expression $q$. To generate an IT expression we keep creating random terms (see Eq.\ref{eq: term}) until either this procedure returns an empty term or we do not have enough budget. Each term is generated by repeatedly drawing a random variable without replacement together with a corresponding random exponent. A dummy variable is inserted into the set to signal when to stop drawing new variables, so initially the process has a probability of $1/(d+1)$ of choosing one of the variables or returning a null term. If it draws a variable in the first turn, the second turn will have a probability of $1/d$ and so on.

\subsection{Crossover}

The crossover procedure will choose two parents through tournament to take part of the process. In the first step, the procedure draws a random point of the first parent to make the recombination. If this point is located at the transformation function $g$, it will generate a child with $g, p$ taken from the first parent and $q$ from the second. If the point is located at the IT expression $p$, it will create a child with $g, q$ from the first parent and a new $p$ as the recombination of both parents. Likewise, if the point is at $q$ the child will have $g, p$ from the first parent and a recombined $q$.


\subsection{Mutation}

The mutation procedure is an uniformly random choice between the set of mutations \emph{insertNode, removeNode, changeVar, changeExponent, changeFunction}. The \emph{insertNode} mutation will either randomly insert a new variable in one of the terms of either $p$ or $q$, or insert a new term in one of those expressions. In the same way, $removeNode$ will either remove a variable or a term. The operators \emph{changeVar, changeExponent, changeFunction} will change one of such elements chosen at random. Notice that in the special case where the number of nodes in the expression exceeds the budget, it will remove the operator \emph{insertNode} from the set.

\section{Experiments}\label{sec:method}

To evaluate the proposed algorithm, we have used the symbolic regression benchmark framework \emph{srbench}~\footnote{\url{https://cavalab.org/srbench/}}. This framework contains $122$ regression datasets with no knowledge of the generating function. This benchmark executes each algorithm $10$ times with a fixed set of random seeds for every dataset. In each run, the framework first splits the dataset with a ratio of $0.75/0.25$ for training and testing, respectivelly, subsampling the training data in case it contains more than $10,000$ samples. After that, it performs a halving grid search~\cite{jamieson2016non} with $5$-fold cross-validation to choose the optimal hyperparameters among $6$ user defined choices. Finally, given the chosen hyperparameters, it executes the algorithm with the whole training data to determine the optimal regression model. It evaluates this model in the test set and stores the approximation error information, running time, number of nodes, and the symbolic model.

SRBench calculates and store the mean squared error (mse), mean absolute error (mae), and coefficent of determination ($R^2$) measurements. The results of each dataset are summarized using the median to minimize the impact of outliers and they are reported with error plots sorted by the median of this summarization. In this work we will focus on the $R^2$ metric following the same analysis in~\cite{2021srbench}.

Currently the \emph{srbench} repository contains results for $21$ regression algorithms, $14$ of which are Symbolic Regression algorithms. From these SR algorithms, $10$ use an evolutionary algorithm as a search procedure. In the next section, we will first report an overall result of the performance of TIR compared to these approaches followed by a more detailed analysis of different aspects of the results. It is important to notice that $62$ of the $122$ datasets are variations of the Friedman benchmark with different numbers of variables, samples, and noise levels. As such, we will also report a separate analysis for the Friedman and non-Friedman sets.

As all the results reported in~\cite{2021srbench} are publicly available at their repository, we only executed the experiments for TIR with the same random seeds so that the results remain comparable. We executed these experiments using Intel DevCloud instance with an Intel(R) Xeon(R) Gold 6128 CPU @ 3.40GHz using only a single thread. The original experiment was executed on an Intel(R) Xeon(R) CPU E5-2690 v4 @ 2.60GHz, so this may create a bias when comparing the execution time.

\subsection{Hyperparameters}

As the rules for this benchmark framework allows only $6$ hyperparameters combinations for the grid search procedure, some of the TIR hyperparameters were fixed after some smaller experiments performed on a subset of the datasets used in~\cite{ITEAforSymbolicRegression}. Table~\ref{tab:fixedpar} shows the choice of hyperparameters used during the \emph{srbench} experiment.

\begin{table}[t]
    \centering
    \caption{Hyperparameters for the TIR algorithm used during the regression benchmark experiments.}
    \begin{tabular}{cc}
        \toprule 
        Parameter &  value \\ \midrule 
        Pop. size &  $1000$ \\ 
        Gens.     & $500$ \\ 
        Cross. prob. & $30\%$ \\ 
        Mut. prob & $70\%$ \\ 
        Transf. functions &  $[id,tanh, sin, cos, log, exp, sqrt]$ \\ 
        Invertible functions     & $[id,atan, tan, tanh, log, exp, sqrt]$ \\ 
        Error measure & $R^2$ \\ 
        budget & $max(5, min(15, \left\lfloor\frac{n\_samples}{10}\right\rfloor)$ \\ 
        $k_{ij}$ range (Eq.~\ref{eq: term}) & $\{(-5, 5), (0, 5), (-2, 2), (0, 2), (-1, 1), (0, 1) \}$ \\
        \bottomrule
    \end{tabular}
    \label{tab:fixedpar}
\end{table}

The budget formula is based on a common rule-of-thumb~\cite{harrell2017regression} that we should have a number of samples of at least $10$ times the number of variables for linear regression. We clip this value in the range $[5, 15]$ so that we do not limit the size of the model to a very small number of terms and neither creates overly long expressions. Regarding the choice of functions set for $g$, note that before the search main loop, the algorithm eliminates from this set any function whose inverse cannot be applied to the training target value. Also notice that we did not use a pre-scaling of the training data and target values unlike some of the algorithms in \emph{srbench}. Finally, during fitness evaluation, the first $90\%$ training samples are used for the ordinary least squares procedure and the last $90\%$ for calculating the fitness of the final model. We chose to have this intersection between the fitting and validation data because some of the data sets contains very few samples, and this was causing overfitted models in some pre-experiments.

\begin{figure*}[t!]
    \centering
    \includegraphics[width=0.8\linewidth]{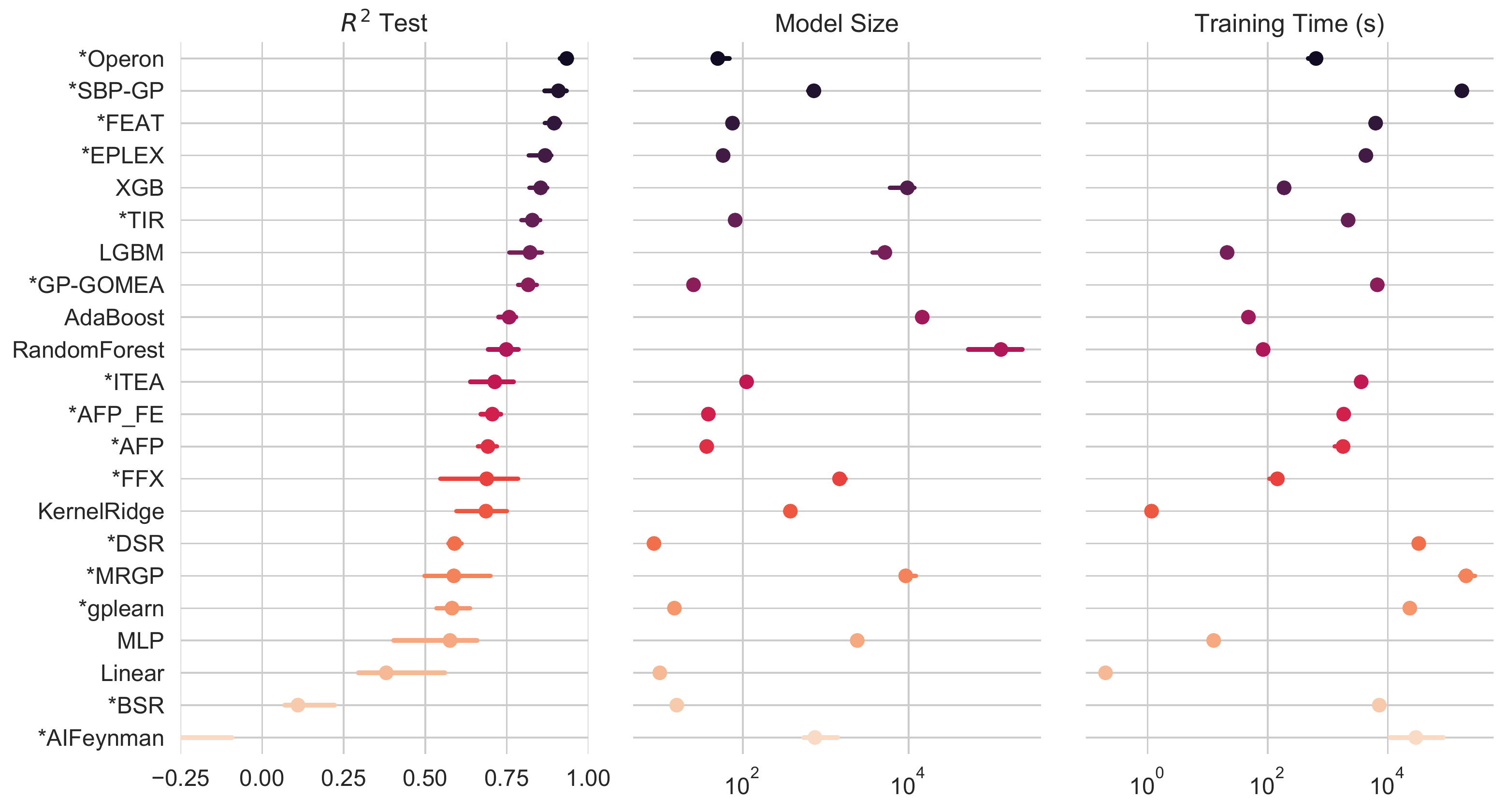}
    \caption{Median of the median results of the \emph{srbench}. Notice that the TIR algorithm was executed on a different machine, so the comparison of the training time may be imprecise.}
    \label{fig:main}
\end{figure*}

\begin{figure}[t]
    \centering
    \includegraphics[width=0.6\linewidth]{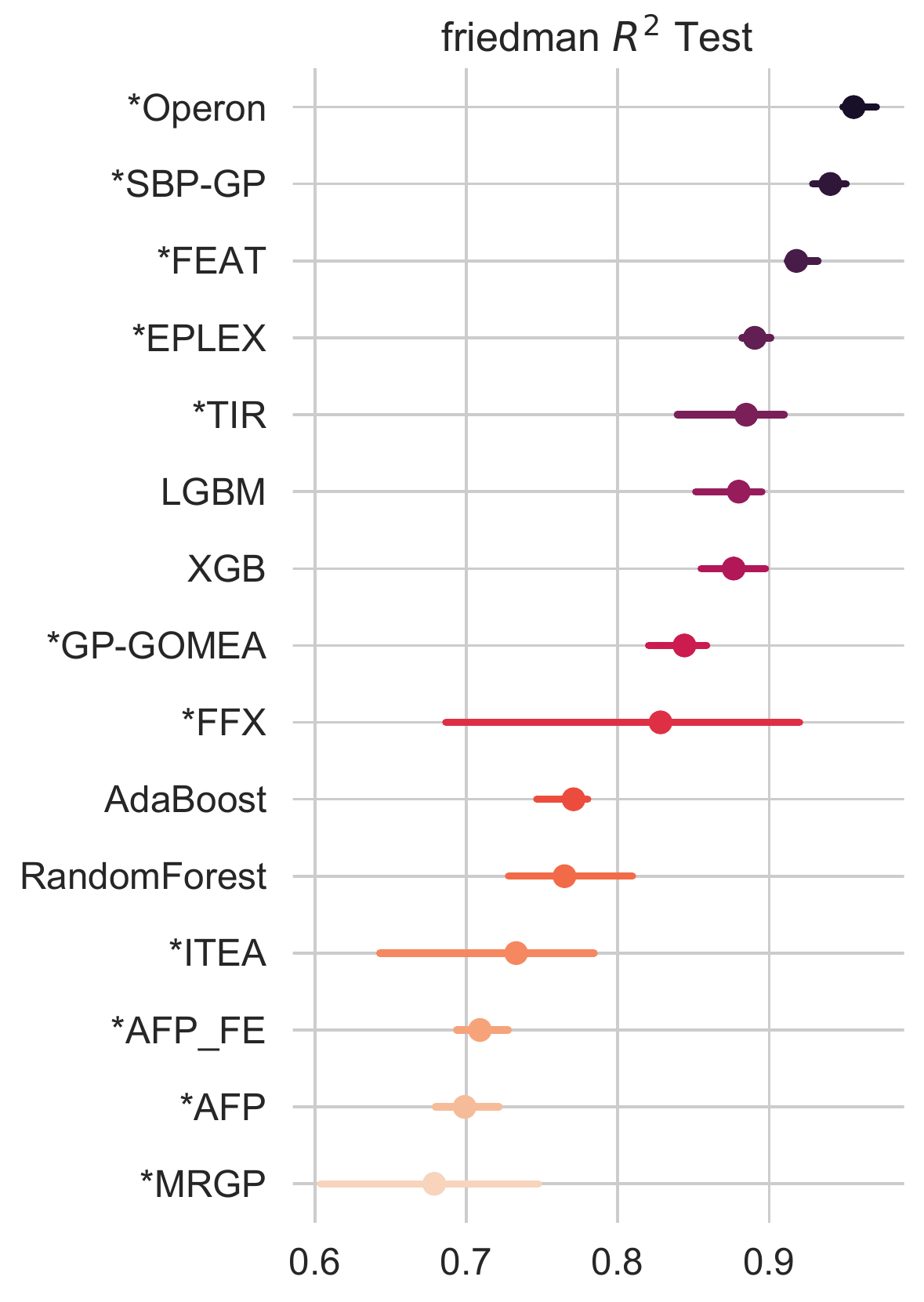}
    \caption{Top $15$ algorithms when considering only the Friedman datasets.}
    \label{fig:fri}
\end{figure}

\begin{figure}[t]
    \centering
    \includegraphics[width=0.6\linewidth]{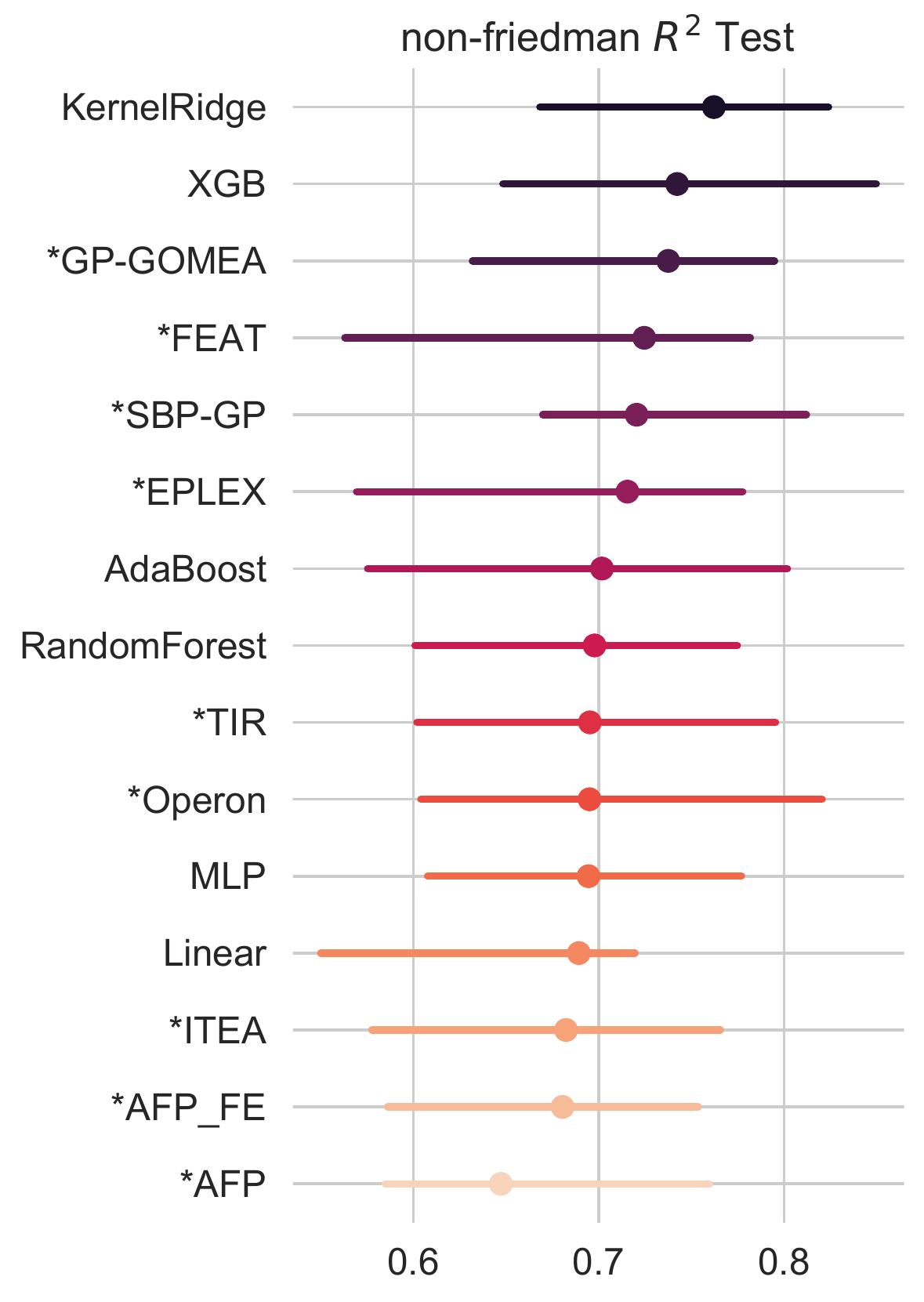}
    \caption{Top $15$ algorithms when considering only the non-Friedman datasets.}
    \label{fig:nonfri}
\end{figure}

\section{Results}\label{sec:results}

In this section we will summarize the obtained results using error bars and critical difference diagrams. For the error bars, we estimate the error using a thousand bootstrap iterations with a confidence interval of $0.95$. The point in this plot represents the median of medians of the $R^2$ calculated over the test sets and the bar represents the range of values that contains the true median with $95\%$ of confidence. The critical difference diagrams uses the Nemenyi test with $\alpha = 0.05$ as a post-hoc test to find the groups of algorithms that presents a significant difference to each other. This test is calculated using the average rank of each algorithm on each dataset using $R^2$ of the test set as the ranking criteria.

\subsection{Initial results}

In Fig.~\ref{fig:main} we can see the overall rank using the median of the medians of the $R^2$ calculated over the test sets of the entire benchmark. In this plot, TIR is ranked $6$th, a five position increase when compared to its predecessor, ITEA. As it turns out, this algorithm now belongs to the set of algorithms with a median in the right end of the $R^2$ spectrum. The error bar indicates that its median is not far from what is reported, ranging from $0.82$ to $0.85$. Comparing with the better ranked approaches, TIR median value is $0.1$ less than Operon, which is ranked first.  Regarding the black-box approaches, TIR presents a very close result to XGB and LGBM ($5$th and $7$th, respectivelly) and better results than other black-box models.

Because of the $budget$ parameter, TIR model sizes are comparable to the top GP approaches, maintaining a median close to $100$ similar to Operon, FEAT and EPLEX, and significantly smaller than SBP-GP and the black-box approaches. Regarding the training time, TIR was a bit faster than ITEA, EPLEX and FEAT, but those were executed in a different machine, thus these values may vary. Even though it used a slower processor, Operon runtime was still faster than TIR. Operon is a heavly optimized GP framework built with computing performance in mind. Besides, TIR also incorporates the ordinary least squares step when evaluating the fitness which dominates the computational costs of this implementation.

As previously mentioned, more than half of the benchmarks are variations of the Friedman benchmark, used to test the innate feature selection capabilities of GP algorithms. As such, the results can be biased toward the algorithm that is more competent in this class of benchmark. In Figs.~\ref{fig:fri}~and~\ref{fig:nonfri}, we show rank of the median of medians of the $R^2$ when considering only the Friedman and non-Friedman datasets, respectively. We can see that, when considering only the Friedman datasets, TIR is ranked $5$th, evidencing that the evolutionary algorithms are more capable of handling the feature selection task in parallel with the model search. When looking at the non-Friedman results, TIR is ranked $9$th. This is due to the increase in rank observed for the black-box approaches. In this specific rank, Kernel Ridge and XGB is ranked first and second, respectivelly. The only GP approaches in the top-$5$ are GP-GOMEA, FEAT and SBP-GP.

\subsection{Penalized fitness for small datasets}

When inspecting the obtained results~\footnote{\url{https://github.com/folivetti/tir/tree/main/srbench/results}}, we noticed that TIR performed worse on smaller datasets due to overfitting when applying ordinary least squares with just a few data points. In this situation, TIR was capable of finding a perfect score for every hyperparameter combination during the grid search and a perfect score on the training data with the optimal parameters, even though it obtained a much worse score in the test set (often a negative $R^2$ value). 
To alleviate this problem, we introduce a penalization term in the fitness function:

\begin{equation*}
    \operatorname{fit}'(x) = \operatorname{fit}(x) - c \cdot \operatorname{size}(x),
\end{equation*}
so that the original fitness value is reduced proportional to the size of the expression. 
As already mentioned, this penalization could not be a part of the grid search since when the issue appears the training $R^2$ score is exactly $1.0$. So, the penalized fitness must be applied according to some rules based on the dataset size. As such, we propose three \emph{ad hoc} rules to verify how much it improves the results, the rules will apply the penalization whenever we have i) a number of samples smaller than $100$ (TIR-samples); ii) the dimension smaller than $6$ (TIR-dim); iii) the product of the number of samples with the dimension smaller than $1,000$ (TIR-points).

\begin{table*}[t]
\caption{Expressions generated by Operon, TIR and penalized TIR for the $192$ vineyard dataset with the random seed $11284$. This is an example chosen to illustrate the benefits, not to be taken as a representative of the average simplicity of either methods.}
\begin{tabular}{@{}llll@{}}
    \toprule
    Algorithm & Expression & $R^2$ \\ \midrule
    Operon & $\frac{\frac{-28643.10 - 30479.64 \cdot x_0}{\sqrt(1 + (4205.37 \cdot x_1)^2} + \frac{21305.05 \cdot x_0 - 36.03 \cdot x_0 \cdot 227.11 \cdot x_1}{\sqrt{1 + (-56.40 - 170.41 \cdot x_0 - 3.58 \cdot x_1)^2}} + 259.47 + 61.52 \cdot x_0}{\sqrt{1 + (-33.91 \cdot x_1 - 339.93 - 161.49 \cdot x_0)^2}}$ & $0.3466$ \\ 
    
    TIR & $\frac{-2.8 + 1.17 \cdot log(x_1^5) - 2.42 \cdot sin(x_1^2 \cdot x_0^5) - 1.10 \cdot sin(x_1^5 \cdot x_0^5) + 9.20 \cdot tanh(x_0) - 1.10 \cdot np.sin(x_0)}{1.0 - 0.09 \cdot sin(x_1^3 \cdot x_0^4) - 0.05 \cdot sin(x_1^2 \cdot x_0)}$ & $0.2621$ \\ 
    
    TIR + penalty & $tan(1.52 - 0.20 \cdot tanh(\frac{1}{x_1 \cdot x_0}))$ & $0.6233$ \\
    \bottomrule
\end{tabular}
\label{tab:exampleexpr}
\end{table*} 

For this next set of experiments, we used a penalization constant of $c = 0.01$. We illustrate the benefits of a penalty function in one of these smaller data sets in Table~\ref{tab:exampleexpr}. In this table we can see the solutions generated by Operon and TIR, both overly long and with a small $R^2$ while a much better solution is obtained with a simple and small expression. We argue that such penalization may benefit other SR algorithms as well.

\begin{figure}[t!]
    \centering
    \includegraphics[width=0.6\linewidth]{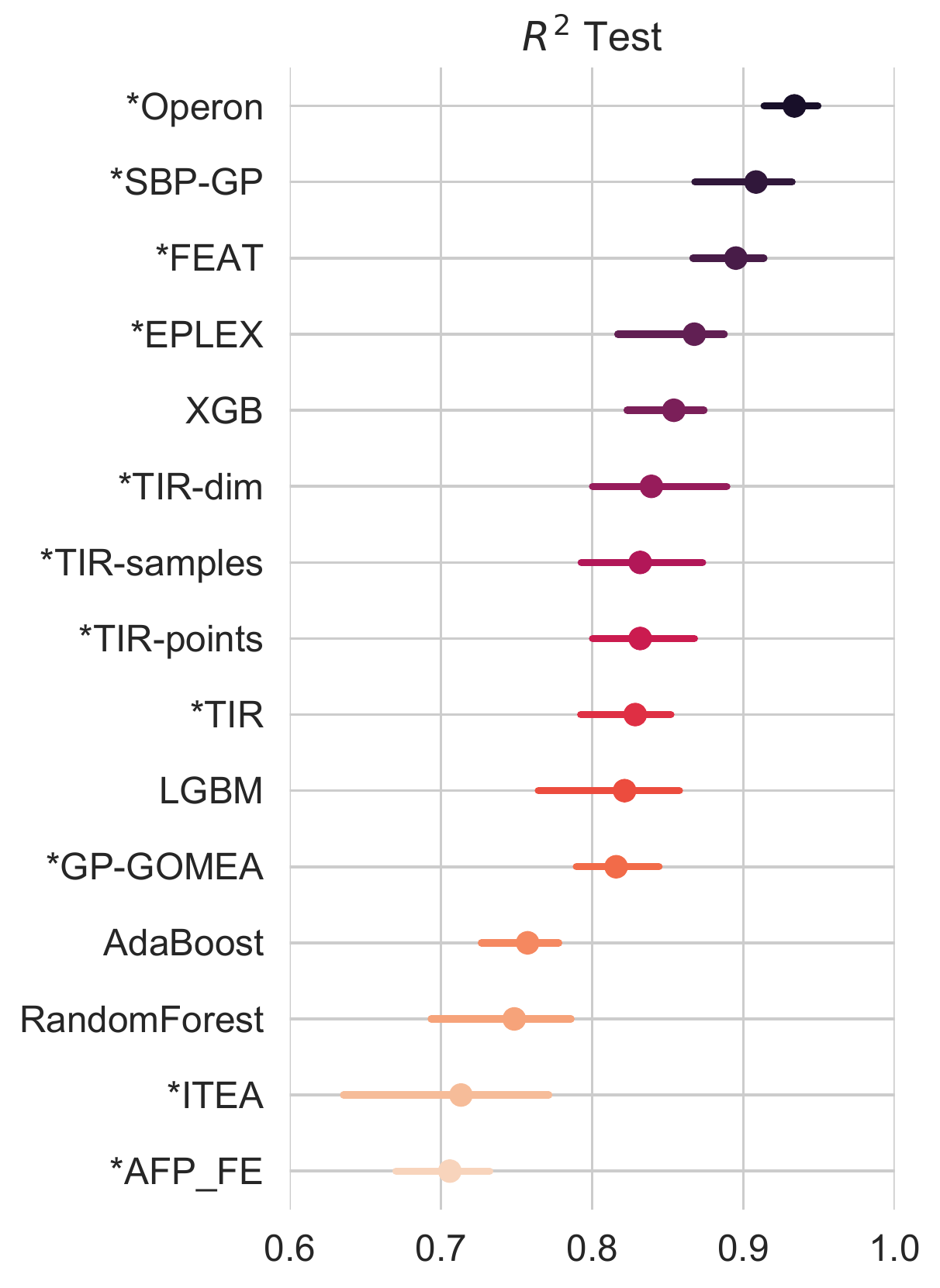}
    \caption{Top $15$ median of medians when considering the three penalization strategies with all of the datasets.}
    \label{fig:penalty}
\end{figure}

\begin{figure}[t!]
    \centering
    \includegraphics[width=0.6\linewidth]{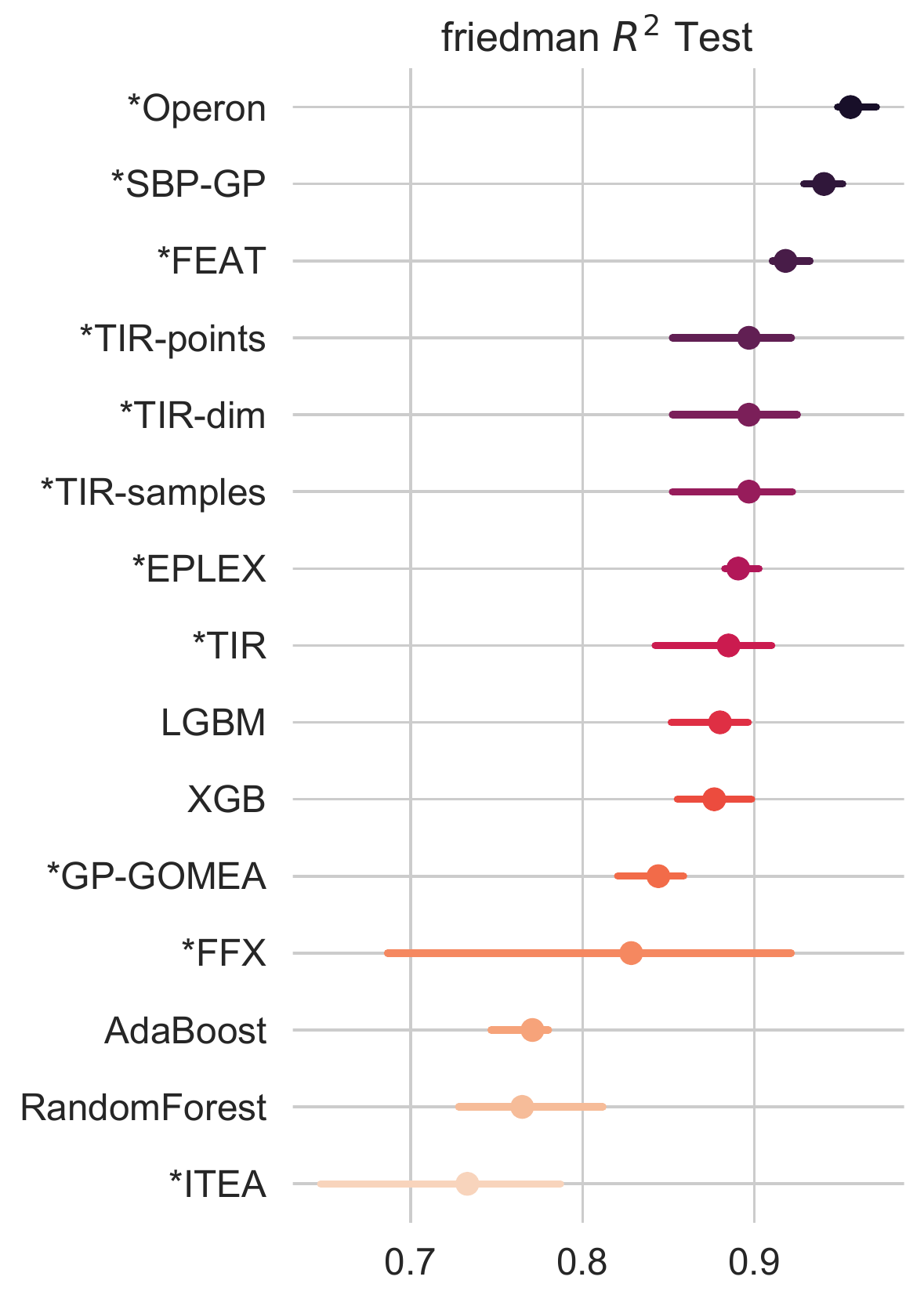}
    \caption{Top $15$ median of medians when considering the three penalization strategies for the Friedman datasets.}
    \label{fig:penalty_fri}
\end{figure}

\begin{figure}[t!]
    \centering
    \includegraphics[width=0.6\linewidth]{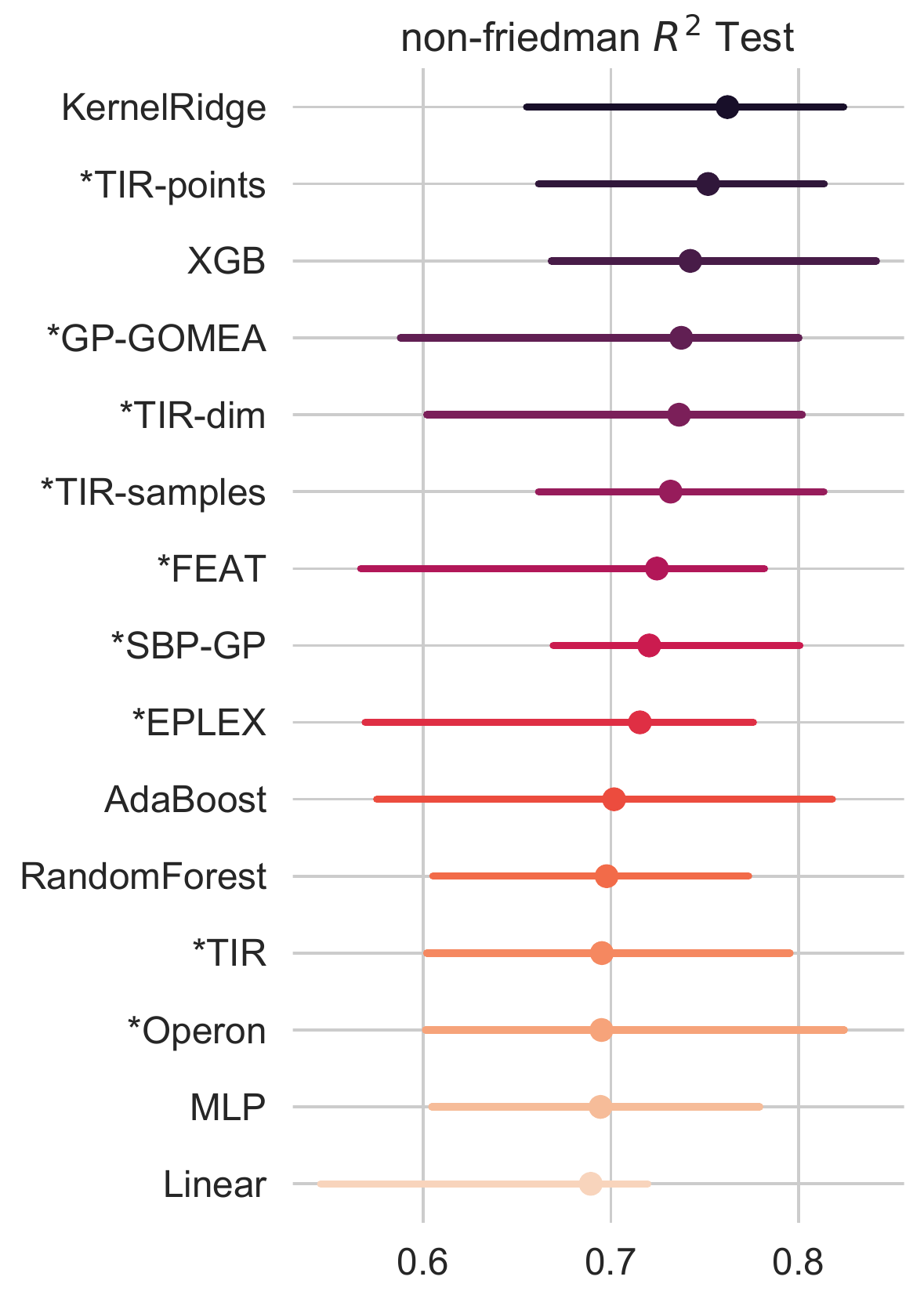}
    \caption{Top $15$ median of medians when considering the three penalization strategies for the non-Friedman datasets.}
    \label{fig:penalty_nonfri}
\end{figure}

Figs.~\ref{fig:penalty},~\ref{fig:penalty_fri},~and~\ref{fig:penalty_nonfri} show the algorithms ranked by the median of medians of the $R^2$ metric for the overall results, Friedman datasets, and non-Friedman datasets, respectively. Since we are using the median of medians to calculate the rank, the overall rank is barely affected since all of the penalization criteria affect less than $20$ datasets. In this first plot we can see that this criteria does not affect the overall performance negatively. When looking only at the Friedman datasets (Fig.~\ref{fig:penalty_fri}), we can see that all of the penalization strategies increase the rank in one position, making TIR the $4$th in rank.

These strategies seem to work particularly well when applying them to the non-Friedman datasets. As we can see in Fig.~\ref{fig:penalty_nonfri}, all of the strategies presented an increase in rank but, particularly the strategy TIR-points, is now ranked $2$nd, in between two black-box approaches, Kernel Ridge and XGB. The confidence interval indicates that the top-$3$ algorithms present similar results.


\subsection{Average rank and critical difference}

\begin{figure}[t]
    \centering
    \includegraphics[width=0.85\linewidth]{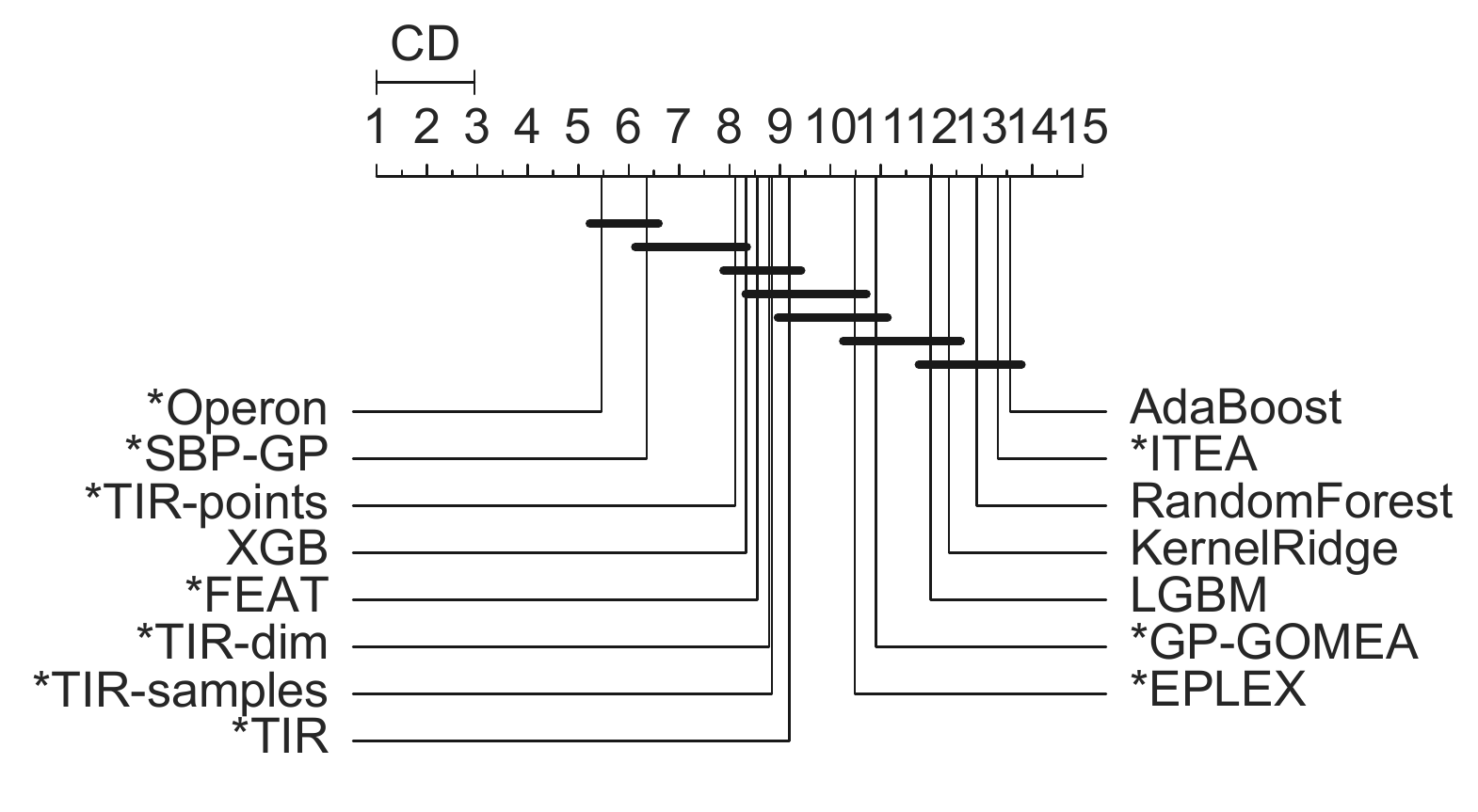}
    \caption{Critical difference diagram of the top $15$ algorithms for the overall results. This plot is computed using the Nemenyi test with $\alpha = 0.05$ calculated over the average rank.}
    \label{fig:cd}
\end{figure}

\begin{figure}[t]
    \centering
    \includegraphics[width=0.85\linewidth]{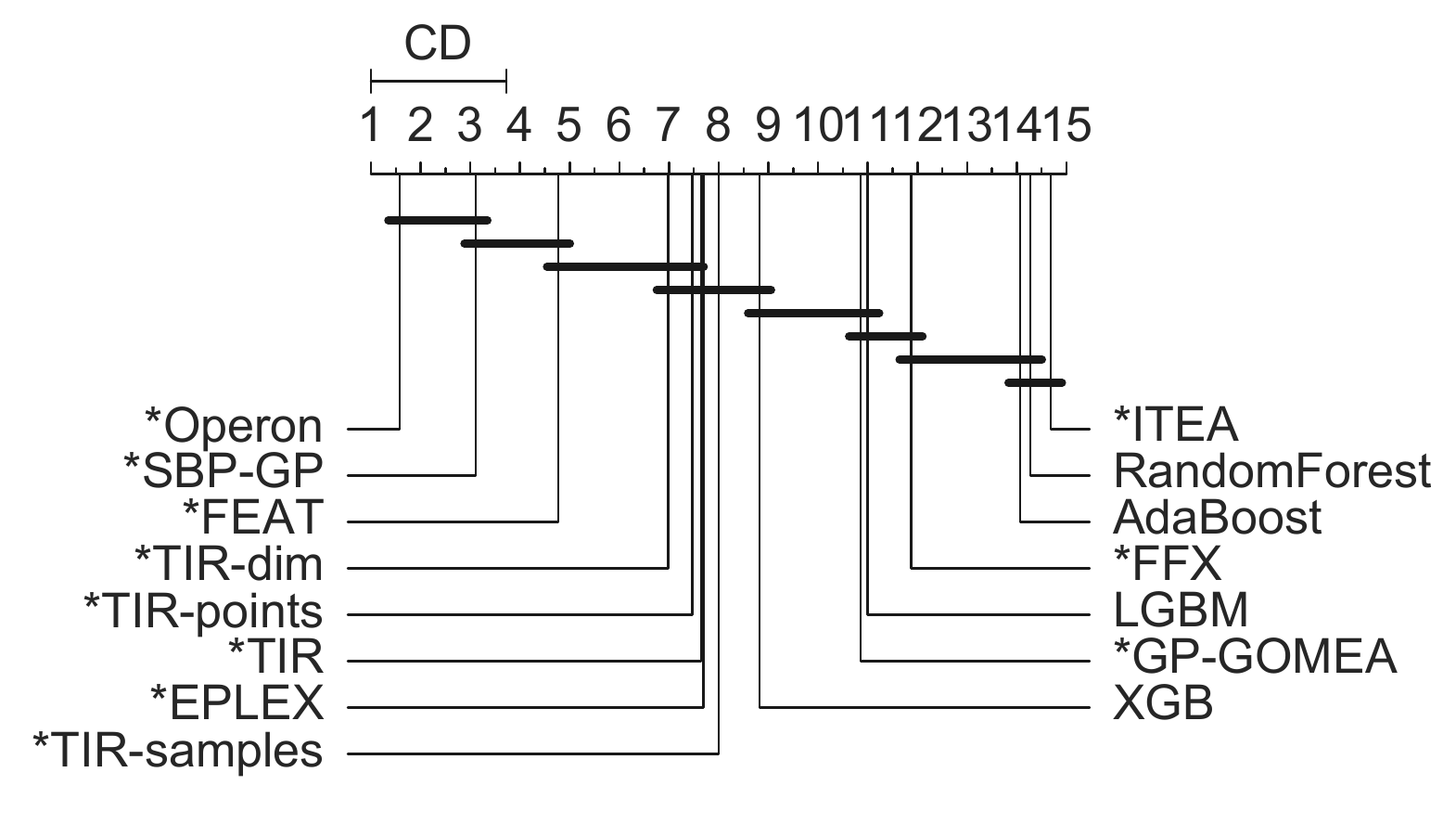}
    \caption{Critical difference diagram of the top $15$ algorithms for the Friedman datasets results. This plot is computed using the Nemenyi test with $\alpha = 0.05$ calculated over the average rank.}
    \label{fig:cdfri}
\end{figure}

\begin{figure}[t]
    \centering
    \includegraphics[width=0.85\linewidth]{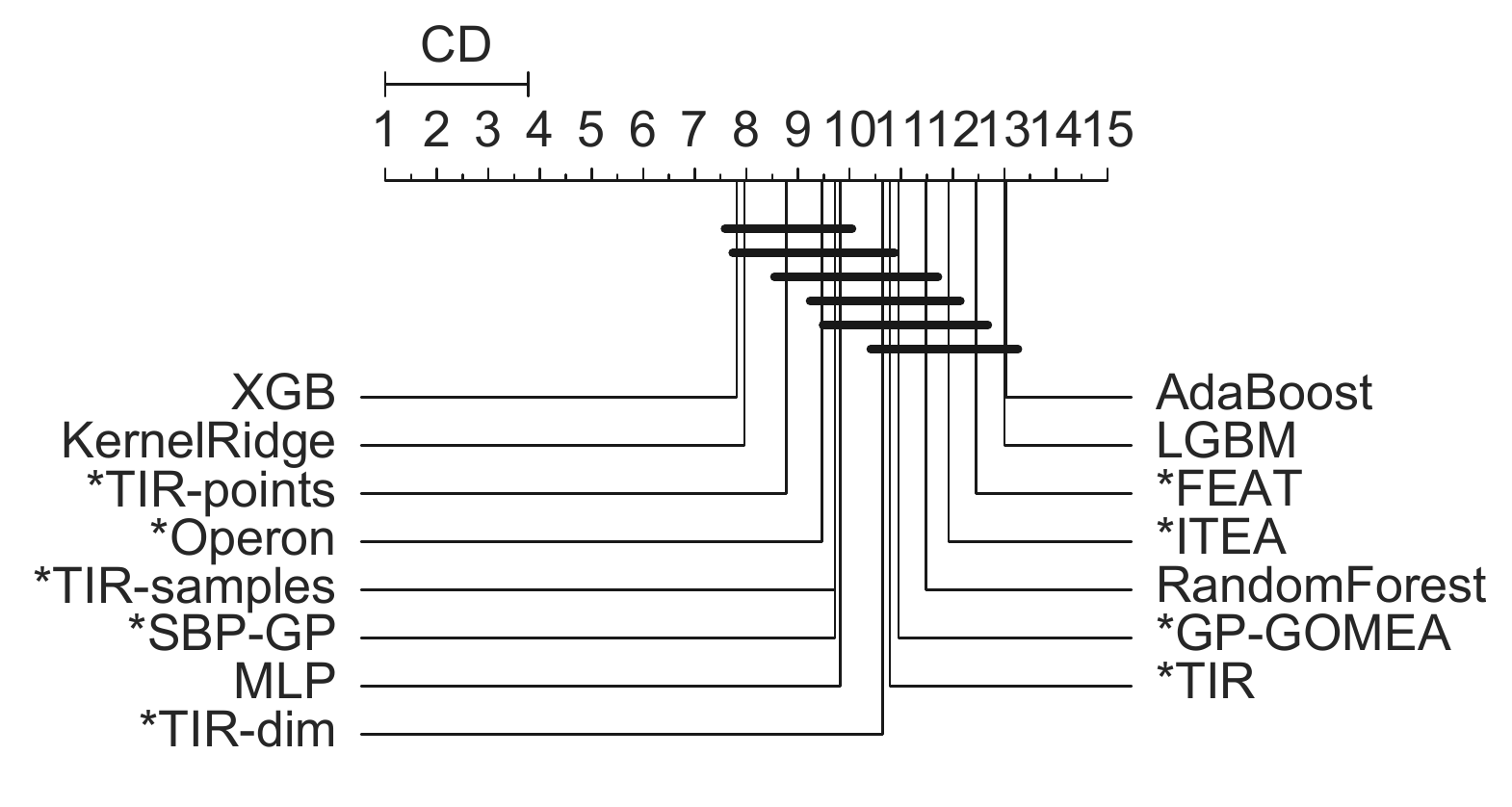}
    \caption{Critical difference diagram of the top $15$ algorithms for the non-Friedman datasets results. This plot is computed using the Nemenyi test with $\alpha = 0.05$ calculated over the average rank.}
    \label{fig:cdnonfri}
\end{figure}

Finally, in Figs~\ref{fig:cd},~\ref{fig:cdfri},~and~\ref{fig:cdnonfri} we show the critical difference diagram using the Nemenyi test with $\alpha = 0.05$ calculated over the average rank. Different from the previous plots, this diagram ranks the algorithms by averaging the rank for each individual problem, thus it depicts a different view from the previous plots.
Considering the entire benchmark we can see that, when looking at the average rank, TIR-points is ranked third with no significant difference with the second in rank and the next five in rank (three of which are TIR variations).
This particular result not only indicates that the median of medians rank only show a partial view of the results, as the rank from the third place forward is changed, but it also shows that there is still insufficient data to draw a conclusive observation of the difference between the third, fourth and fifth approaches (excluding TIR variations). 

Looking at the Friedman datasets, TIR-dim and TIR-points are ranked fourth and fifth, respectively. They both have no significant difference with the third place, FEAT, but there is a significant difference to the first and second place. When looking at the non-Friedman datasets, TIR-points is ranked third without any significant difference with the first and second place, both black-box models. Again, the horizontal bar indicates that further experiments are necessary, with more datasets, to draw conclusive observations.


\section{Conclusion}\label{sec:conclusion}

In this paper we propose a novel representation for Symbolic Regression, called Transformation-Interaction-Rational (TIR) as an extension of Interaction-Transformation (IT) representation. 
This representation extends the IT representation as a rational of two expressions composed with a unary function. With this new representation, many more function forms can be represented while prohibiting some complicated forms such as those with function chaining (except for the target transformation). Together with the new representation, we presented a framework for the evolutionary search of TIR expressions composed of supporting libraries aimed at improving the search process. Another benefit of this representation, when coupled with modal arithmetic, is the guarantee of generating only valid expressions without the need of protected operators.

We tested this new representation following a simple evolutionary search procedure on a thorough benchmark specifically crafted to evaluate symbolic regression approaches. Overall the results showed a significant improvement over its predecessor while maintaining a good rank position in different scenarios.

Particularly when testing penalization strategies for small data--sets, we observed a performance rivaling black-box approaches (top-$3$) for a certain class of datasets. For the Friedman datasets, the proposed approach still mantained a good position in rank, having comparable results against other Symbolic Regression approaches.

For the next steps, we will investigate other strategies for dealing with overfitting and the use of adaptive hyperparameters to avoid the burden of having to apply a grid search strategy. Additionally, we will consider the inclusion of inner adjustable parameters in this representation similar to what is done in GP NLS/Operon. While this may add an extra cost to the search process, it can help to decrease the approximation error in some situations.


  

%% file: floats/texts/ITRepresentationExamples.tex
\begin{tabular}{@{}llll@{}}
    \toprule\midrule
    Equation & IT & TIR \\ \midrule
    $x_0 (x_1 + x_2 x_3 sin(x_4))$  & -- & -- \\\midrule
    $\frac{x_0}{\sqrt{1 - \frac{x_1^2}{x_2^2}}}$ &  -- & $\sqrt{\frac{x_0^2}{1 - x_1^2 x_2^{-2}}}$  \\ \midrule
    $\frac{1}{1 + x_0^{-4}} + \frac{1}{1 + x_1^{-4}}$ &  -- & $\frac{x_0^4 +  x_1^4 + 2 x_0^4 x_1^4}{1 + x_0^4 x_1^4 + x_0^4 + x_1^4}$  \\ \midrule
    $\sin{\left(\frac{x_0^2}{x_1}\right) + \frac{1}{2} e^{\frac{x_1}{x_0}}}$ & $\sin{(x_0^2  x_1^{-1})} + 0.5  exp{(x_0  x_1^{-1})}$ & $ \sin{(x_0^2  x_1^{-1})} + 0.5  exp{(x_0  x_1^{-1})}$ \\\bottomrule
\end{tabular}

